\pdfoutput=1
\PassOptionsToPackage{table}{xcolor}
\documentclass[11pt]{article}

\usepackage[preprint]{acl}

\usepackage{times}
\usepackage{latexsym}

\usepackage[T1]{fontenc}
\usepackage[utf8]{inputenc}
\usepackage{xcolor}
\usepackage{comment,pbox,multirow,adjustbox}
\usepackage{pifont,graphics,textcomp,array,float,caption,subcaption}
\usepackage{arabtex,utf8}
\usepackage{enumitem}
\usepackage{CJKutf8}
\usepackage{pgfplots,pgfplotstable}
\usepackage{cleveref}
\usepackage{microtype}

\usepackage{inconsolata}

\usepackage{graphicx}
\usepackage{acro}
\acsetup{
  uppercase/list
}
\pgfplotsset{compat=1.8}
\usepgfplotslibrary{statistics}

\newcommand\CS{CSW}
\newcommand\lexDict{{LEX$_{Dict}$}}
\newcommand\lexRand{{LEX$_{Rand}$}}
\newcommand\LexPred{{LEX$_{Pred}$}}

\newcommand\ecRand{{EC$_{Rand}$}}
\newcommand\ecSPF{{EC$_{SPF}$}}
\newcommand\mlRand{{ML$_{Rand}$}}
\newcommand\mlSPF{{ML$_{SPF}$}}
\newcommand\ec{{EC}}
\newcommand\ml{{ML}}
\newcommand\BT{{BT}}

\definecolor{DarkGreen}{RGB}{1,150,32}
\definecolor{DarkRed}{RGB}{1,150,32}

\title{The Impact of Code-switched Synthetic Data Quality is Task Dependent:  Insights from MT and ASR}

\author{Injy Hamed,$^{1}$ Ngoc Thang Vu,$^{2}$ Nizar Habash$^{1,3}$\\
  $^1$MBZUAI, $^2$University of Stuttgart, $^3$New York University Abu Dhabi\\
  \texttt{injy.hamed@mbzuai.ac.ae}, \texttt{thang.vu@ims.uni-stuttgart.de}
  \\\texttt{nizar.habash@nyu.edu}
  }

\begin{document}
\maketitle
\begin{abstract}
Code-switching, the act of alternating between languages, emerged as a prevalent global phenomenon that 
needs to be addressed for building user-friendly language technologies. A main bottleneck in this pursuit is data scarcity, motivating research in the direction of code-switched data augmentation. However, current literature lacks comprehensive studies that enable us to understand the relation between the quality of synthetic data and improvements on NLP tasks. We extend previous research conducted in this direction on machine translation (MT) with results on automatic speech recognition (ASR) and cascaded speech translation (ST) to test generalizability of findings. Our experiments involve a wide range of augmentation techniques, covering lexical replacements, linguistic theories, and back-translation. Based on the results of MT, ASR, and ST, we draw conclusions and insights regarding the efficacy of various augmentation techniques and the impact of quality on performance.
\end{list} % known arabtex bug

\end{abstract}

\setcode{utf8}

\section{Introduction}
%With technology becoming an integral part of people's lives, it has become ubiquitous to enable technology to embrace the diverse linguistic needs and preferences of users, rather than forcing users to operate within the limitations of technology. 
%Given that more than half of the world's population is multilingual speakers \citep{grosjean2013psycholinguistics}, 
Code-switching (CSW) is a worldwide phenomenon, involving the alternation between multiple languages in the same discourse.\footnote{For survey papers on CSW in NLP, we refer the readers to the following papers: \citet{sitaram2019survey,dougruoz2021survey,winata2023decades,hamed2025survey}.} 
%Despite its prevalence, language technologies are not yet capable of processing it as seamlessly as monolingual data 
Despite the need to process it effectively, language technologies still fall short when handling code-switched input compared to monolingual data \citep{dougruoz2021survey}, where the lack of CSW resources is a main challenge. CSW data augmentation has thus been gaining attention as a workaround for alleviating this issue. %the stark lack of CSW data. %, improving the performance of NLP models. 
Furthermore, the need for language technologies to not only process, but also generate CSW in human-computer interaction has been highlighted by researchers \cite{bawa2020multilingual} for the aim of building tools that cater to the needs and preferences of multilingual communities.
%CSW generation is also important for enabling language technologies to communicate with humans in CSW  thus building inclusive and user-friendly tools that cater to the needs and preferences of multilingual communities.

%Despite the considerable amounts of efforts in this direction, 
While considerable amount of research has been conducted on CSW data augmentation,  
we still lack comprehensive studies covering multiple augmentation techniques, human and extrinsic evaluations, and multiple NLP tasks. Such studies are needed to draw conclusions with regards to the improvements achieved by the different augmentation techniques on NLP tasks, the quality of the generated augmentations, and the relation between both; quality and improvements.% on NLP tasks.
%between quality of synthetic data and the improvements achieved on downstream tasks.
%most of the previous studies on this front have focused on one augmentation technique without exploring others, or reported results using only one type of word alignments configuration, or evaluated effectiveness of augmentation on only one downstream task.

Several studies evaluate the effectiveness of different augmentation techniques extrinsically, however lack human evaluations assessing the quality of generations \cite{WMW+18,WMW+19,LV20,GVS21}. Other studies include human evaluations, however, do not report results on downstream tasks \cite{PC21,kuwanto2024linguistics}. Few studies involve both extrinsic and human evaluations. The study by \citet{hussein2023textual} involved two augmentation approaches, however, their effectiveness was only reported in the scope of speech recognition. \citet{HHA+22} presented a study that is diverse in terms of extrinsic evaluations, covering MT, AST, and ST,  however, the augmentation approaches were limited to lexical replacements only. Finally, \citet{hamed2023CSAug} presented a comprehensive study covering multiple augmentation techniques, however, the extrinsic evaluation only covered the task of MT. % the extrinsic evaluation was only conducted on MT. 
%The effectiveness of the techniques was measured on MT. 

Given current literature, we cannot draw strong conclusions with regards to the effectiveness of the different techniques across different NLP tasks, as well as the relation between quality and improvements achieved on downstream tasks. 
%it is still unclear whether the findings generalize to other NLP tasks as each of the previous papers have their own setup, and are either limited in the number of augmentation approaches, or the NLP tasks covered, or both. 
In this paper, we aim at extending current literature with findings based on a more comprehensive setup in terms of investigated augmentation techniques and NLP tasks. 
%In this paper, we extend this line of research, where
To achieve that, we build %on the same experimental setup of their work, 
on our experimental setup in \citet{hamed2023CSAug}, being the most comprehensive study in terms of augmentation approaches. We report 
%, following the same setup of data,  experiments, and model, and 
%reporting
results on ASR and cascaded ST, covering a wider range of NLP tasks investigated within the same experimental setup. %With this, we aim at allowing researchers to draw more solid conclusions on the relation between the quality of CSW text generation and its effectiveness on downstream tasks.

Our contributions are as follows:
\begin{itemize}
    \item Following our previous experimental setup in \citet{hamed2023CSAug}, we report new results on ASR and ST. This allows us to make comparisons and draw conclusions based on five variations of augmentation approaches (covering linguistic-based approaches, lexical replacements, and back-translation) and three downstream tasks (MT, ASR, and ST).
    \item We present a discussion on the relation between the quality of generations and their effectiveness on NLP tasks in light of the results on ASR and ST as well as previous MT results. Our results show that with regards to the effectiveness of the techniques, some approaches are consistent in their performance across tasks, while others are task-dependent. Moreover, we find that the relation between data quality and NLP improvements, while confirmed for MT, does not hold for ASR. 
    \item We explore and discuss other factors, besides quality of generations, that may affect results, including data diversity and task complexity.
\end{itemize}

The paper is organized as follows. Section~\ref{sec:rw} discusses related work. In Section~\ref{sec:data_aug_techniques}, we provide an overview on the augmentation techniques included in the study. Section~\ref{sec:exp_setup} is dedicated to the experimental setup. In Section~\ref{sec:results}, we present the ASR and ST results, as well as the correlations between quality of augmentations and ASR improvements. Finally, in Section~\ref{sec:discussion}, we provide further insights, discussing the possible impact of other factors. %multiple factors that may impact results.
%\begin{itemize}
    %\item importance of handling CSW
    %\item bottleneck: data scarcity -> data augmentation
    %\item research gap: a lot of work, many reserchers putting effort into generating natural CSW sentences. In the context of using this data for improving NLP systems, does it really matter?
    %\item Lacking comparative studies that include human evaluation and multiple approaches to study the relation between quality and performance. Only aware of 1 paper that presented such a study for MT, do the results hold to other NLP tasks?
    %\item We replicate that experiment for ASR and extend their results and ours to report results on a cascaded ST system.
%\end{itemize}

\section{Related Work}
\label{sec:rw}
%Previous research on \CS~data augmentation has addressed multiple NLP tasks, including sentiment analysis (https://arxiv.org/pdf/2211.07628), POS tagging, NER, and X. The majority of 
The majority of previous research on \CS~data augmentation has addressed language modeling (LM), primarily for ASR. Various techniques have been investigated based on heuristics \citep{SWY+11,VLW+12,KAT+21}, linguistic theories \citep{PBC+18,LYL19,hussein2023textual}, 
MT \citep{TKJ21}, and generative models \citep{WMW+18,WMW+19,LV20} 
including large language models (LLMs) \cite{hu2023improving,alharbi2024leveraging}. 
%CCL18,SZY+19,MLJ+19
MT has received less attention, 
where techniques mainly involved lexical replacements \citep{AGE+21,GVS21,XY21} 
%MLJ+19,SZY+19,AGE+21,GVS21,XY21
and few efforts investigated back-translation \citep{KFC+21} and linguistic theories \cite{hamed2023CSAug}. 

With regards to studies conducting human evaluations without experimental results on downstream tasks, \citet{PC21} compared lexical replacements and linguistic-based approaches, where higher human preference was observed for the latter approach. Recently, \citet{kuwanto2024linguistics} investigated the use of the Equivalence Constraint theory \citep{Pop80} when prompting LLMs by providing information on words that should be code-switched, showing slight improvements.
%MLJ+19,SZY+19

With regards to studies comparing different augmentation techniques through human evaluations as well as extrinsically, 
%\citet{kuwanto2024linguistics} relied on LLMs to transform a monolingual sentence into its CSW equivalence, while investigating the effectiveness of  using a relaxed version of the Equivalence Constraint linguistic theory to inform the LLM on words that should be code-switched. 
%%\citet{PC21} compared lexical replacements and linguistic-based approaches based on human evaluation. While higher human preference was observed for generations using linguistic-based approaches, the study did not include extrinsic evaluations on downstream tasks, and thus the effectiveness on NLP models was not measured. %, and therefore their effectiveness was not measured.
%assess improvements on 
\citet{hussein2023textual} compared random lexical replacements versus utilizing the Equivalence Constraint theory through human evaluation and ASR results. While the linguistic-based approach was found to be superior in the human evaluation, 
%the superiority of the linguistic-based approach was reported in human evaluation
it was outperformed by random lexical replacements in language modeling and speech recognition. In \citet{HHA+22}, we compared different approaches for lexical replacements. While the authors provide a comprehensive study, including human evaluation and results on MT, ASR, and ST tasks, the study is focused on experimental considerations for lexical replacements and does not include other augmentation approaches. 
In \citet{hamed2023CSAug}, we presented a comprehensive study covering multiple augmentation techniques, including linguistic-based approaches, lexical replacements, and back-translation. The study involved extrinsic evaluation on MT task in addition to human evaluation assessing the naturalness of the generations across techniques. % the extrinsic evaluation was only conducted on MT. 
%The effectiveness of the techniques was measured on MT. 
A positive correlation was reported between the naturalness scores achieved by the different techniques and improvements on MT. However, given that the study is only focused on MT, 
it is still unclear whether the findings generalize to other NLP tasks.
%they only only focus on lexical replacements. % without comparison to other techniques.
%\begin{itemize}
%    \item focus on ASR
%    \item Point out the scarcity in work on CSW ST, especially in data augmentation
%    \item focus on comparative studies
%\end{itemize}
In this paper, we work towards filling the current research gap in comparative studies, extending literature with further findings in the area of CSW data augmentation.

\section{Data Augmentation Techniques}
\label{sec:data_aug_techniques}
We cover the same techniques and setup as in \citet{hamed2023CSAug}, where Arabic-English parallel sentences are utilized to generate CSW Arabic-English sentences using the approaches below.
\subsection{Lexical Replacements}
\paragraph{Dictionary Replacement (\lexDict):} $X$\% random Arabic words on the source side are replaced with their English gloss entries. The gloss entries are obtained using MADAMIRA \cite{PAD+14}, a system that performs morphological analysis and disambiguation for Arabic. $X$ is set to 19 based on the frequency found in naturally occurring CSW data \cite{hamed2022arzenST}. 

\paragraph{Aligned with Random CSW Points\ (\lexRand):} 
$X$\% Arabic words on the source side are replaced with their counterpart English words on the target side based on alignments obtained using Giza++ \cite{och03:asc}, as specified in \citet{hamed2023CSAug}. $X$ is also set to 19.
%We augment the Arabic-to-English parallel sentences by randomly picking 19\% source-target aligned words and replacing the source words with their counterpart words on the target side. 

\paragraph{Aligned with Predicted CSW Points (\LexPred):} 
%A CSW predictive model \cite{AGE+21,HHA+22} 
Instead of randomly choosing the words on the target side to be injected into the source side, a CSW predictive model is leveraged, where the model identifies the words on the target side that would be plausible CSW words on the source side. The CSW predictive model from \citet{HHA+22} 
is utilized for this task. The model is trained using ArzEn-ST corpus \cite{hamed2022arzenST}, containing CSW Arabic-English sentences and their English translations. In order to train the CSW predictive model, a matching algorithm was developed to tag the words on the target side that match the code-switched words on the source side. An mBERT model is then fine-tuned on this binary classification task, where given an English sentence, the model identifies which words are probable to be present in the CSW corresponding sentence.
Afterwards, similar to the previous augmentation approach, target-to-source replacements are performed using alignments to inject these words into the source side sentence. 
%Similar to the previous approach, we perform target-to-source replacements; however, the choice of words on the target side to be inserted into the source side is based on . The predictive model is trained to 

\subsection{Linguistic Theories (\ec~and \ml)}
The GCM tool \citep{RSG+21} is utilized to obtain CSW generations following two linguistic theories: Equivalence Constraint (EC) \citep{Pop80} and Matrix Language Frame (MLF) \citep{myers1997duelling}. %The two theories represent two dominant approaches to intra-sentential (word-level) \CS; alternational  and insertional. 
%For both theories, we use the GCM tool \citep{RSG+21}. 
The tool provides two approaches for sampling across the multiple generations it provides;  random and Switch Point Fraction (SPF). In SPF sampling, the generations are ranked based on their SPF \citep{PBC+18} distribution compared to a reference SPF distribution. % and the generations are chosen accordingly. %SPF is calculated as the number of switch points divided by the total number of (language-dependent) tokens in the sentence.
%\footnote{The current version of the tool provides an implementation for Switch Point (SP) which does not account for the number of tokens in the sentence. Therefore, we implement our own code for ranking based on SPF.} 
The reference SPF (0.22) is calculated based on %the \CS~sentences in ArzEn-ST train set 
natural CSW data \citep{hamed2022arzenST}. 
Similar to the previous approaches, one generation is sampled per sentence for both sampling variants. We refer to the variants as \ecRand, \ecSPF, \mlRand, and \mlSPF.

\subsection{Back-translation (\BT)}
A BT model \cite{hamed2023CSAug} is trained to translate English into CSW Arabic-English. The model is utilized to 
translate the target side of the Arabic-to-English parallel sentences to CSW sentences. The model is trained as a Transformer model using Fairseq \cite{OEB+19} by utilizing the Arabic-English parallel corpora discussed in Section \ref{sec:data} in addition to ArzEn-ST corpus, where the approach is outlined in \citet{hamed2023CSAug}.

\section{Experimental Setup}
\label{sec:exp_setup}
\subsection{Data}
\label{sec:data}
%We use ArzEn-ST \cite{hamed2022arzenST} corpus  We also utilize Callhome \citep{GKA+97} and MGB-3 \citep{AVR17} for monolingual Egyptian Arabic data, Librispeech \citep{PCP+15} for English data, and MGB-2 \citep{ABG+16} for MSA data. 
In this Section, we specify the datasets used in (1) generating the augmentations and (2) training and evaluating the ASR systems. 

For augmentation, we use the synthetic data generated in \citet{hamed2023CSAug}. % to allow for comparison across tasks, and evaluation of the techniques in the cascaded ST task. %the cascaded ST task. 
The generations are obtained by augmenting 309$k$ Arabic-English parallel sentences collected from the following corpora: % for generating augmentations:
Callhome Egyptian Arabic-English Speech Translation
Corpus \cite{GKA+97,LDC2002T38,LDC2002T39,KCC+14}, LDC2012T09 \cite{ZMD+12}, 
LDC2017T07 \cite{LDC2017T07}, 
LDC2019T01 \cite{LDC2019T01}, 
LDC2021T15 \cite{LDC2021T15}, and MADAR \cite{BHS18}. Using the approaches outlined in Section~\ref{sec:data_aug_techniques}, these monolingual parallel sentences are augmented into CSW Arabic-English sentences.

For ASR, we utilize ArzEn-ST,  %\citep{hamed2022arzenST}, 
which is a CSW Arabic-English speech translation corpus. 
The corpus contains naturally occurring speech having frequent CSW \cite{hamed2020arzen} along with its Arabic and English translations. The corpus is used in training, development and testing. ArzEn-ST %corpus 
train, dev, and test sets contain 
3.3k, 1.4k, and 1.4k sentences (containing 2.2k, 0.9k, and 0.9k CSW sentences). 
For training, we also utilize Callhome \citep{GKA+97} and MGB-3 \citep{AVR17} for Egyptian Arabic data, in addition to 5 hours from each of Librispeech \citep{PCP+15} for English data, and MGB-2 \citep{ABG+16} for Modern Standard Arabic (MSA) data. %The preprocessing steps are outlined in Appendix~\ref{appendix:data_processing}.
We perform Arabic Alif/Ya normalization, remove punctuation and corpus-specific annotations, and lower-case English words.

%For monolingual Egyptian Arabic data, we use Callhome \citep{GKA+97} containing 19.4 hours of telephone conversations and MGB-3 \citep{AVR17} consisting of 16 hours gathered of Egyptian YouTube videos. For the MGB-3 corpus, we excluded utterances having overlapping speech, thus ended up with a subset of 9-hours. For monolingual English speech data, we utilize Librispeech \citep{PCP+15}, containing 1,000 hours of read audiobooks. For MSA, we utilize MGB-2 \citep{ABG+16}, containing 1,137 hours gathered from Aljazeera Arabic TV programs covering news domain. For MGB-2 and Librispeech, we used subsets of the corpora, following (cite ASR paper). 

\subsection{ASR Model}
We use joint CTC/attention based end-to-end ASR systems using ESPnet \citep{WHK+18}. We apply SpecAugment \citep{PCZ+19} and set the CTC/attention weight %$(\lambda)$ is set 
to 0.3. The encoder and decoder consist of 12 and 6 Transformer blocks with 4 heads, feed-forward inner dimension 2048 and attention dimension 256 \citep{KSWDON+19}. We use RNNLM consisting of 1 LSTM layer with 1000 hidden units trained for 20 epochs. For decoding, we set the beam size to 20 and CTC weight to 0.2. The LM is trained on the transcriptions of the ASR corpora, in addition to the synthetic CSW data in case of data augmentation experiments.
\subsection{ST Model}
We evaluate the effectiveness of the augmentation techniques on a cascaded ST task. We utilize our ASR models and the MT models from \citet{hamed2023CSAug}, where we train Transformer models using Fairseq. 
We report results on ArzEn-ST test set.
 
%\subsection{Augmentation Models}
%%\subsection{Zero-shot and Non-zero-shot Settings}
%We cover both settings:% in our ASR evaluation:
%%\begin{itemize}
%%    \item Zero-shot setting: given the scarcity of CSW resources, we mimic the case of the lack of CSW corpora. We train a baseline model, ASR\_BL$_{Mono}$, using the monolingual speech corpora for Egyptian Arabic, English, and MSA only. Data augmentation is performed %for the 309$k$ monolingual Arabic to English parallel sentences 
%%    using the techniques that do not require \CS~parallel corpora: \lexDict, \lexRand, {\sc EC}, and {\sc ML}. The augmented CSW data along with the monolingual speech corpora transcriptions are used for LM rescoring.
%%    \item Non-zero-shot setting: this setting allows the use of CSW corpora. The baseline model, ASR\_BL$_{All}$, is trained on the monolingual speech corpora in addition to ArzEn-ST. For augmentation, all techniques are applied. %included in the comparison.
%%\end{itemize}

\section{Results}
\label{sec:results}
We present ASR and ST results and discuss the relation between naturalness scores of the generations and improvements on ASR. For ASR, the full results are presented in Table~\ref{table:ASR_extrinsic_eval_results}. We present WER and CER on ArzEn-ST test set, for all sentences as well as CSW sentences
only. We also report perplexity (PPL), out-of-vocabulary (OOV) rates, and the number of generations per technique. For ST, the full results are provided in Table~\ref{table:ST_extrinsic_eval_results}, showing BLEU \cite{papineni2002bleu}, chrF, chrF++ \cite{popovic2017chrf++}, and BERTScore (F1) \cite{ZKW+19}, reported on all ArzEn-ST test set and the CSW sentences only.
We provide the statistical significance for both tasks in Appendix~\ref{sec:appendix_full_results}. The analysis in this section is based on the results on ArzEn-ST test set CSW sentences, using WER and chrF++, as CSW is our main concern. For easier comparison of results across ASR and MT, we also briefly discuss previous results obtained on MT. 
%\subsection{ASR Baselines}
%We develop two baselines for zero-shot and non-zero-shot setups. We train the ASR model on the monolingual speech corpora outlined in Section \ref{sec:data}, where the LM is trained on the transcriptions. This baseline, referred to as ASR\_BL$_{Mono}$, is used for the zero-shot setting. We also train another model, ASR\_BL$_{All}$, where we utilize the monolingual data in addition to ArzEn-ST train set. This baseline is used for the non-zero-shot setting.

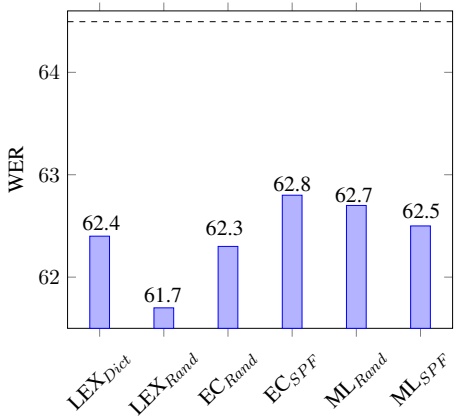
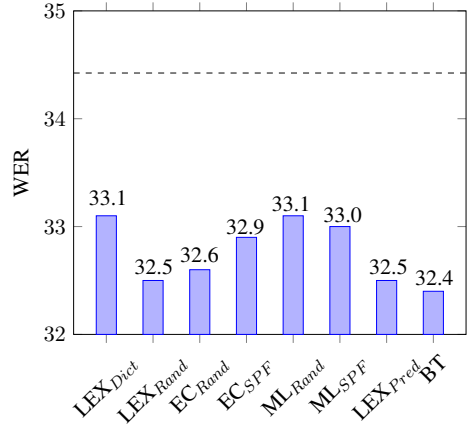
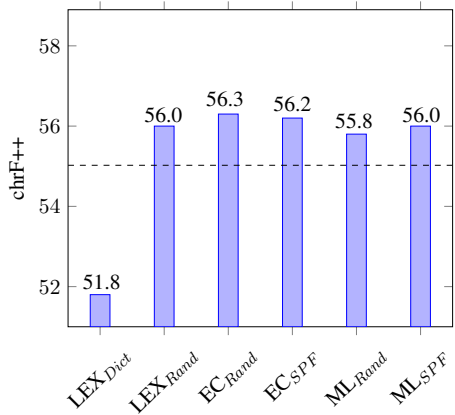
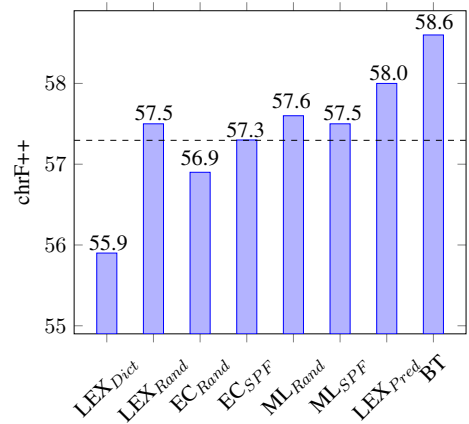
\begin{figure*}[t]
    \centering
    \begin{subfigure}[b]{0.45\textwidth}
        \centering
        %    \begin{subfigure}{}
        %\centering
        \resizebox{0.85\columnwidth}{!}{%
        \begin{tikzpicture}
        \begin{axis}[ybar, symbolic x coords={\lexDict, \lexRand, \ecRand, \ecSPF, \mlRand, \mlSPF}, ylabel={WER},
        %grid=both,
        %major grid style={gray},
        ymin=61.5,
        ymax=64.6,
        xticklabel style={rotate=45}
        ]
        \addplot+ coordinates {(\lexDict, 62.4) (\lexRand, 61.7) (\ecRand, 62.3) (\ecSPF, 62.8) (\mlRand, 62.7) (\mlSPF, 62.5)};
        \end{axis}
        \node (n1) at (0.6,1.9) {62.4};
        \node (n2) at (1.7,0.6) {61.7};
        \node (n3) at (2.8,1.8) {62.3};
        \node (n4) at (4.0,2.6) {62.8};
        \node (n5) at (5.1,2.4) {62.7};
        \node (n6) at (6.3,2.1) {62.5};
        \draw (n1) (n2) (n3) (n4) (n5) (n6);
        \draw [dashed] (0,5.5) -- (6.85,5.5);
        %\draw [dashed] (0,3.41) -- (6.85,3.41);
        \end{tikzpicture}
        }
        %    \end{subfigure}
            \caption{
            %The effectiveness of the augmentation techniques on ASR in zero-shot (upper) and non-zero-shot (lower) settings. We show the WER on ArzEn-ST test set \CS~sentences. The solid line represents ASR\_BL$_{Mono}$ and the dashed line represents ASR\_BL$_{All}$.
            The WER achieved by the ASR models %on ArzEn-ST test set \CS~sentences %by applying the augmentation techniques 
            in zero-shot setting. The dashed line represents ASR\_BL$_{Mono}$.
            }
            \label{fig:ASR_zero-shot}
    \end{subfigure}
    \hfill
    \begin{subfigure}[b]{0.45\textwidth}
    \centering
    %\label{fig:ASR_zero-shot}
    %    \end{subfigure}
        %~~~~~~~~~~~~~~~~~~~~~~~~~~~~~~~~~~~~~
     %   \begin{subfigure}{}
    \resizebox{0.85\columnwidth}{!}{%
    \begin{tikzpicture}
    \begin{axis}[
        ybar, symbolic x coords={\lexDict, \lexRand, \ecRand, \ecSPF, \mlRand, \mlSPF, \LexPred, \BT}, 
        xtick=data, %to force all x-axis labels to be displayed
        ylabel={WER},
        ymin=32,
        ymax=35,
        xticklabel style={rotate=45}
    ]
    \addplot+ coordinates {(\lexDict, 33.1) (\lexRand, 32.5) (\ecRand, 32.6) (\ecSPF, 32.9) (\mlRand, 33.1) (\mlSPF, 33.0) (\LexPred, 32.5) (\BT, 32.4)};
    \end{axis}
    \node (n1) at (0.6,2.4) {33.1};
    \node (n2) at (1.4,1.2) {32.5};
    \node (n3) at (2.2,1.4) {32.6};
    \node (n4) at (3.0,1.9) {32.9};
    \node (n5) at (3.8,2.3) {33.1};
    \node (n6) at (4.7,2.1) {33.0};
    \node (n7) at (5.5,1.2) {32.5};
    \node (n8) at (6.3,1.0) {32.4};
    \draw (n1) (n2) (n3) (n4) (n5) (n6) (n7) (n8);]
    %baseline:57.3
    \draw [dashed] (0,4.6) -- (6.85,4.6);
    %\draw (0,9) -- (6.85,9);
    \end{tikzpicture}
    }
    %    \end{subfigure}
        \caption{
        The WER achieved by the ASR models %on ArzEn-ST test set \CS~sentences 
        in non-zero-shot setting. The dashed line represents ASR\_BL$_{All}$.
        }
        \label{fig:ASR_non-zero-shot}
    \end{subfigure}
    \vskip\baselineskip
    \begin{subfigure}[b]{0.45\textwidth}
        \centering
        \resizebox{0.85\columnwidth}{!}{%
        \begin{tikzpicture}
        \begin{axis}[ybar, symbolic x coords={\lexDict, \lexRand, \ecRand, \ecSPF, \mlRand, \mlSPF}, ylabel={chrF++},
        %grid=both,
        %major grid style={gray},
        ymin=51,
        ymax=58.9,
        xticklabel style={rotate=45}
        ]
        \addplot+ coordinates {(\lexDict, 51.8) (\lexRand, 56) (\ecRand, 56.3) (\ecSPF, 56.2) (\mlRand, 55.8) (\mlSPF, 56)};
        \end{axis}
        \node (n1) at (0.6,0.8) {51.8};
        \node (n2) at (1.7,3.8) {56.0};
        \node (n3) at (2.8,4.1) {56.3};
        \node (n4) at (4.0,4.0) {56.2};
        \node (n5) at (5.1,3.7) {55.8};
        \node (n6) at (6.3,3.8) {56.0};
        \draw (n1) (n2) (n3) (n4) (n5) (n6);
        \draw [dashed] (0,2.9) -- (6.85,2.9);
        %\draw [dashed] (0,3.41) -- (6.85,3.41);
        \end{tikzpicture}
        }
        \caption{The chrF++ scores achieved by the MT models %on ArzEn-ST test set \CS~sentences 
        in zero-shot setting. The dashed line represents MT\_BL$_{Mono}$.
        }
        \label{fig:zero-shot}
    \end{subfigure}
    \hfill
    \begin{subfigure}[b]{0.45\textwidth}
        \centering
        \resizebox{0.85\columnwidth}{!}{%
        \begin{tikzpicture}
        \begin{axis}[
            ybar, symbolic x coords={\lexDict, \lexRand, \ecRand, \ecSPF, \mlRand, \mlSPF, \LexPred, \BT}, 
            xtick=data, %to force all x-axis labels to be displayed
            ylabel={chrF++},
            ymin=54.9,
            ymax=58.9,
            xticklabel style={rotate=45}
        ]
        \addplot+ coordinates {(\lexDict, 55.9) (\lexRand, 57.5) (\ecRand, 56.9) (\ecSPF, 57.3) (\mlRand, 57.6) (\mlSPF, 57.5) (\LexPred, 58) (\BT, 58.6)};
        \end{axis}
        \node (n1) at (0.6,1.6) {55.9};
        \node (n2) at (1.4,3.9) {57.5};
        \node (n3) at (2.2,3.1) {56.9};
        \node (n4) at (3.0,3.6) {57.3};
        \node (n5) at (3.8,4.1) {57.6};
        \node (n6) at (4.7,3.9) {57.5};
        \node (n7) at (5.5,4.6) {58.0};
        \node (n8) at (6.3,5.5) {58.6};
        \draw (n1) (n2) (n3) (n4) (n5) (n6) (n7) (n8);]
        %baseline:57.3
        \draw [dashed] (0,3.41) -- (6.85,3.41);
        %\draw (0,0.15) -- (6.85,0.15);
        \end{tikzpicture}
        }
        \caption{The chrF++ scores achieved by the MT models %on ArzEn-ST test set \CS~sentences 
        in non-zero-shot setting. The dashed line represents MT\_BL$_{All}$.
        }
        \label{fig:non-zero-shot}
    \end{subfigure}
    \caption{ASR and MT results on ArzEn-ST test set \CS~sentences in zero-shot and non-zero-shot settings.}
    \label{fig:asr_mt_results}
\end{figure*}

\begin{table*}[t]
\centering
\begin{tabular}{| l | l | r | r  r | c  c | c  c |}
\cline{6-9}
\multicolumn{5}{c|}{}&\multicolumn{2}{c|}{\textbf{All Test Sentences}} &\multicolumn{2}{c|}{\textbf{\textbf{CSW Test Sentences}}} \\\hline
\multicolumn{1}{|c|}{\textbf{Exp}}&\multicolumn{1}{c|}{\textbf{Model}}& \multicolumn{1}{c|}{$|$\textbf{Train}$|$} & \textbf{PPL} &\textbf{OOV} & \textbf{WER} &\textbf{CER} & \textbf{WER} &\textbf{CER}\\
%\cline{2-12}
\hline
%\hline
\multicolumn{9}{|c|}{\textbf{Baselines}}\\\hline
B1&ASR\_BL$_{Mono}$ &27,449&687.7&10.57&62.1&38.5&64.5&41.1\\	
B2&ASR\_BL$_{All}$ &30,793&\textbf{415.1}&\textbf{5.57}&\textbf{34.7}&\textbf{20.0}&\textbf{34.4}&\textbf{20.0}\\
\hline\hline
\multicolumn{9}{|c|}{\textbf{Zero-shot Experiments (ASR\_BL$_{Mono}$+Augmentations)}}\\\hline
A1&+{\lexDict} &267,093&396.2&6.62& 60.0&37.0&62.4&39.6\\
A2&+{\lexRand} &220,101&\textbf{364.6}&\textbf{5.70}&\textbf{59.5}&\textbf{36.9}&\textbf{61.7}&\textbf{39.4}\\
A3&+\ecRand &169,549&460.0&6.23&60.2&37.2&62.3&39.6\\
A4&+\ecSPF &169,549&438.8&6.25&60.6&37.4&62.8&39.9\\
A5&+\mlRand &125,681&455.3&6.37&60.5&37.3&62.7&39.8\\
A6&+\mlSPF &125,681&460.9&6.36&60.4&37.4&62.5&39.9\\
\hline\hline
\multicolumn{9}{|c|}{\textbf{Non-zero-shot Experiments (ASR\_BL$_{All}$+Augmentations)}}\\\hline
A7&+{\lexDict}&270,437&318.6&4.16&33.3&19.3&33.1&19.3\\
A8&+{\lexRand}&223,445&274.1&\textbf{\underline{3.88}}&\textbf{32.9}&18.9&32.5&18.8\\
A9&+{\LexPred}&143,735&\textbf{\underline{270.4}}&\textbf{\underline{3.88}}&33.0&18.9&32.5&18.8\\
A10&+\ecRand&172,893&301.0&3.95&33.1&18.9&32.6&18.8\\
A11&+\ecSPF&172,893&309.7&3.93&33.4&19.1&32.9&19.0\\
A12&+\mlRand&129,025&313.7&4.11&33.7&19.3&33.1&19.2\\
A13&+\mlSPF&129,025&297.4&4.09&33.5&19.2&33.0&19.0\\
A14&+\BT&181,868&275.3&3.96&\textbf{\underline{32.9}}&\textbf{\underline{18.8}}&\textbf{\underline{32.4}}&\textbf{\underline{18.7}}\\
\hline\hline
\multicolumn{9}{|c|}{\textbf{Constrained Experiments (ASR\_BL$_{All}$+Constrained[Augmentations])}}\\\hline
A15&+{\lexDict} &55,636&410.2&4.57&34.3&19.7&33.8&19.6\\
A16&+{\lexRand} &55,636&384.8&\textbf{4.39}&34.0&19.5&33.4&19.4\\
A17&+{\LexPred} &55,636&385.4&4.42&34.2&19.5&33.7&19.5\\
A18&+\ecRand &55,636&394.5&4.50&34.2&19.6&33.6&19.5\\
A19&+\ecSPF &55,636&446.2&4.48&34.6&19.7&34.0&19.6\\
A20&+\mlRand &55,636&435.5&4.54&34.6&19.8&34.2&19.8\\
A21&+\mlSPF &55,636&416.1&4.54&34.6&19.7&34.1&19.6\\
A22&+\BT &55,636&\textbf{361.9}&4.41&\textbf{33.7}&\textbf{19.3}&\textbf{33.2}&\textbf{19.2}\\
\hline
\end{tabular}
\caption{ %ASR results
%We report ASR results using WER and CER on ArzEn-ST test set, for all sentences and \CS~sentences only as well as PPL and OOV on all sentences of ArzEn-ST test set.
We report ASR results using WER and CER on ArzEn-ST test set, for all sentences as well as \CS~sentences only. We also report PPL and OOV on all sentences of ArzEn-ST test set. We report the results of the baselines, zero-shot and non-zero-shot settings and well as the constrained settings. Given the varying amounts of generations produced by each technique, we also report the number of sentences used in training each model. The best performing models in each setting are bolded. The overall best performing model is underlined.
%In Table \ref{table:stat_sig}, we show the statistical significance between the models in zero-shot and non-zero shot settings.
} 
\label{table:ASR_extrinsic_eval_results}
\end{table*}

\subsection{ASR Results}
\label{sec:ASR_results}
We report results on the following two settings:
\begin{itemize}
    \item Zero-shot setting: given the scarcity of CSW resources, we mimic the case of the lack of CSW corpora. We train a baseline model, ASR\_BL$_{Mono}$, using the monolingual speech corpora for Egyptian Arabic, English, and MSA only. Data augmentation is performed %for the 309$k$ monolingual Arabic to English parallel sentences 
    using the techniques that do not require \CS~parallel corpora: \lexDict, \lexRand, {\sc EC}, and {\sc ML}. The augmented CSW data along with the monolingual speech corpora transcriptions are used for LM rescoring.
    \item Non-zero-shot setting: this setting allows the use of CSW corpora. The baseline model, ASR\_BL$_{All}$, is trained on the monolingual speech corpora in addition to ArzEn-ST. For augmentation, all techniques are applied. %included in the comparison.
\end{itemize}
We present WER results on ArzEn-ST test set
CSW sentences in Figures~\ref{fig:ASR_zero-shot} and ~\ref{fig:ASR_non-zero-shot}. The baseline models, ASR\_BL$_{Mono}$ and ASR\_BL$_{All}$, achieve 64.5\% and 34.4\% WER, respectively. %For the zero-shot setting, we observe that \lexDict~provides comparable performance to linguistic theories. Among the linguistic theories, the best performance is achieved by \ecRand. Overall, the highest improvement is achieved by \lexRand, with an absolute WER reduction of 2.8\%  over ASR\_BL$_{Mono}$.
For the zero-shot setting, %we observe that 
among the linguistic theories, the best performance is achieved by \ecRand. With regards to lexical replacements, \lexDict~provides comparable performance to linguistic theories and \lexRand~provides highest overall improvements, achieving absolute WER reduction of 2.8\%  over ASR\_BL$_{Mono}$.

For the non-zero-shot setting, %%all augmentation approaches bring improvements over ASR\_BL$_{All}$. 
%This is unlike the MT results, where only \LexPred and \BT outperformed the baseline. Similar to MT, 
the best result is achieved by \BT, achieving 2.0\% absolute WER improvements over ASR\_BL$_{All}$. By checking statistical significance, we find that \lexRand, \LexPred, and \ecRand~provide equal performance to \BT. This is followed by the other linguistic variants and \lexDict. It should be noted that \lexRand~proves to be a strong approach for ASR across both settings, while requiring no linguistic knowledge nor \CS~data. %This was also found in the study by 
This is in-line with the results of \citet{hussein2023textual}, where the superiority of random lexical replacements was demonstrated over the use of the Equivalence Constraint theory for ASR. 

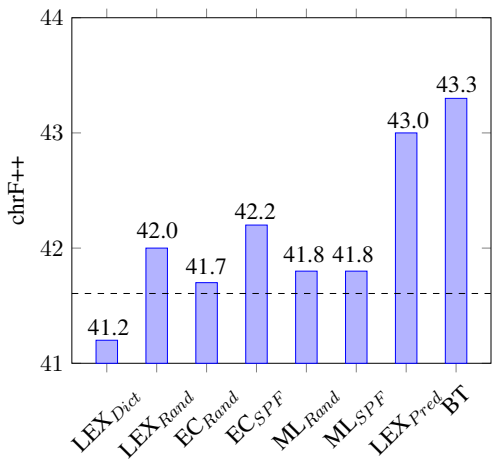
\begin{figure}[t]
\centering
\resizebox{0.85\columnwidth}{!}{%
\begin{tikzpicture}
\begin{axis}[
    ybar, symbolic x coords={\lexDict, \lexRand, \ecRand, \ecSPF, \mlRand, \mlSPF, \LexPred, \BT}, 
    xtick=data, %to force all x-axis labels to be displayed
    ylabel={chrF++},
    ymin=41,
    ymax=44,
    xticklabel style={rotate=45}
]
\addplot+ coordinates {(\lexDict, 41.2) (\lexRand, 42.0) (\ecRand, 41.7) (\ecSPF, 42.2) (\mlRand, 41.8) (\mlSPF, 41.8) (\LexPred, 43.0) (\BT, 43.3)};
\end{axis}
\node (n1) at (0.6,0.6) {41.2};
\node (n2) at (1.4,2.2) {42.0};
\node (n3) at (2.2,1.6) {41.7};
\node (n4) at (3.0,2.5) {42.2};
\node (n5) at (3.8,1.8) {41.8};
\node (n6) at (4.6,1.8) {41.8};
\node (n7) at (5.5,4.0) {43.0};
\node (n8) at (6.3,4.6) {43.3};
\draw (n1) (n2) (n3) (n4) (n5) (n6) (n7) (n8);]
%baseline:57.3
\draw [dashed] (0,1.15) -- (6.85,1.15);
%\draw (0,9) -- (6.85,9);
\end{tikzpicture}
}
\caption{%ST results
The chrF++ scores achieved in ST on ArzEn-ST test set \CS~sentences in non-zero-shot setting. 
The dashed line represents the baseline model ST\_BL$_{All}$.
%ChrF++ scores for ST on ArzEn-ST test set \CS~sentences. The dashed line represents ST\_BL$_{All}$.
}
\label{fig:ST_non-zero-shot}
\end{figure}

\begin{figure}[t]
\centering
\resizebox{0.85\columnwidth}{!}{%
\begin{tikzpicture}
\begin{axis}[
    ybar, symbolic x coords={\lexDict, \lexRand, \ecRand, \ecSPF, \mlRand, \mlSPF, \LexPred, \BT}, 
    xtick=data, %to force all x-axis labels to be displayed
    ylabel={\%Sentences},
    ymin=0,
    ymax=100,
    xticklabel style={rotate=45}
]
\addplot+ coordinates {(\lexDict, 16.0) (\lexRand, 47.3) (\ecRand, 48.6) (\ecSPF, 50.0) (\mlRand, 60.7) (\mlSPF, 45.6) (\LexPred, 66.6) (\BT, 80.0)};
\end{axis}
\node (n1) at (0.5,1.2) {16.0};
\node (n2) at (1.4,2.9) {47.3};
\node (n3) at (2.2,3.0) {48.6};
\node (n4) at (3.0,3.1) {50.0};
\node (n5) at (3.8,3.7) {60.7};
\node (n6) at (4.6,2.9) {45.6};
\node (n7) at (5.5,4.0) {66.6};
\node (n8) at (6.3,4.8) {80.0};
\draw (n1) (n2) (n3) (n4) (n5) (n6) (n7) (n8);
%\draw (0,9) -- (6.85,9);
\end{tikzpicture}
}
\caption{%Human evaluation
The human evaluation scores as obtained from \cite{hamed2023CSAug}, showing the percentage of augmentations perceived as natural per technique.
}
\label{fig:human_eval}
\end{figure}
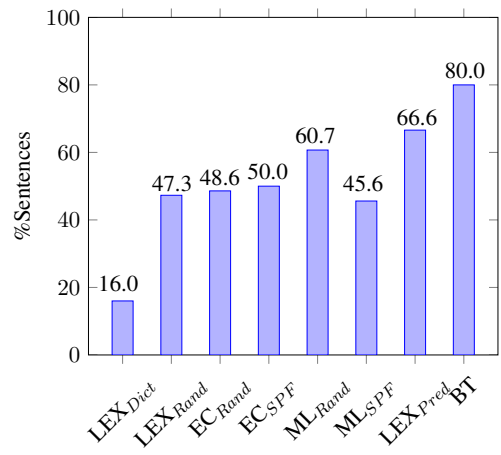

\subsection{MT Results}
We include MT results from \citet{hamed2023CSAug} in Figure~\ref{fig:asr_mt_results}. 
Similar to ASR, MT results cover zero-shot and non-zero-shot settings, with their respective baselines; MT\_BL$_{Mono}$ and MT\_BL$_{All}$. In case of the zero-shot setting, the MT models are trained on the Arabic-English parallel corpora outlined in Section~\ref{sec:data}, in addition to augmentations from the respective approaches. In case of the non-zero-shot setting, the training data of the MT models additionally included ArzEn-ST corpus.
For a full discussion on MT experimental setup and results, we refer the readers to \citet{hamed2023CSAug}.
Across both settings, \lexDict~ degrades MT performance over baselines. We also report that linguistic-based approaches and \lexRand~perform equally well, however, they are unable to achieve significant improvements over the baseline in the non-zero-shot setting. \BT~and \LexPred~ show superiority, achieving +1.3 and +0.7 chrF++ points over the baseline, respectively.

\subsection{ST Results}
\label{sec:ST_results}
%For each pipeline (baseline or augmentation technique), we pass the ASR hypotheses to the MT system to obtain the translations. 
We present the chrF++ scores  on ArzEn-ST test set \CS~sentences for the non-zero-shot setting in Figure \ref{fig:ST_non-zero-shot}. %The full results and statistical significance are provided in Appendix~\ref{sec:appendix_full_results}. 
The baseline, ST\_BL$_{All}$, achieves 41.6 chrF++ points. 
%Using the ASR and MT systems trained in the non-zero-shot setting, we report results for cascaded ST systems where for each pipeline (baseline or augmentation technique) we pass the ASR hypotheses to the MT system to obtain the translations. We present the chrF++ scores in Figure \ref{fig:ST_non-zero-shot} (full results in Table~\ref{table:ST_extrinsic_eval_results}). The baseline in this setup (ST\_BL$_{All}$) utilizes MT\_BL$_{All}$ and ASR\_BL$_{All}$ models, achieving 41.6 chrF++ points on the \CS~sentences in ArzEn-ST test set.
We observe that \lexDict~does not outperform the baseline, where its overall performance on the ST task is affected by the low MT results. 
%low performance on MT. 
Among the linguistic theories, \ecSPF~performs best, and is the only variant that outperforms the baseline, providing similar performance to \lexRand. The best performance is achieved by \BT~followed by \LexPred, achieving improvements of +1.7 and +1.4 chrF++ points over ST\_BL$_{All}$. %, respectively.

\begin{table*}[t]
\centering
\begin{tabular}{| l | l | c  c  c  c  | c  c  c  c |}
\cline{3-10}
\multicolumn{2}{c|}{} &\multicolumn{4}{c|}{\textbf{All Test Sentences}} &\multicolumn{4}{c|}{\textbf{\textbf{CSW Test Sentences}}} \\\hline
\multicolumn{1}{|c|}{\textbf{Exp}}&\multicolumn{1}{c|}{\textbf{Model}}& \textbf{BLEU} &\textbf{chrF} & \textbf{chrF++} &\textbf{BertScore} & \textbf{BLEU} &\textbf{chrF} & \textbf{chrF++} &\textbf{BertScore}\\
%\cline{2-12}
\hline
\hline
\multicolumn{10}{|c|}{\textbf{Baseline}}\\\hline
&+ST\_BL$_{All}$ &15.9&42.2&40.3&0.335&16.4&43.7&41.6&0.318\\\hline\hline
\multicolumn{10}{|c|}{\textbf{Non-zero-shot Experiments %(MT\_BL$_{All}$/ASR\_BL$_{All}$+Augmentations)
}}\\\hline
A7&+{\lexDict} &15.7&42.1&40.2&0.343&16.1&43.2&41.2&0.322\\
A8&+{\lexRand} &15.9&42.7&40.7&0.347&16.5&44.1&42.0&0.329\\
A9&+{\LexPred} &\textbf{17.3}&43.5&41.7&\textbf{0.351}&\textbf{17.9}&44.9&43.0&0.335\\
A10&+\ecRand &15.7&42.5&40.5&0.343&16.1&43.9&41.7&0.324\\
A11&+\ecSPF &16.5&42.8&40.9&0.348&17.1&44.2&42.2&0.334\\
A12&+\mlRand &16.0&42.6&40.6&0.342&16.4&43.9&41.8&0.323\\
A13&+\mlSPF &16.0&42.6&40.6&0.346&16.5&44.0&41.8&0.330\\
A14&+\BT    &16.9&\textbf{43.7}&\textbf{41.8}&0.349&17.7&\textbf{45.4}&\textbf{43.3}&\textbf{0.337}\\\hline
\end{tabular}
\caption{
We report ST results using BLEU, chrF, chrF++, and BertScore (F1) on ArzEn-ST test set, for all sentences as well as \CS~sentences only. We report the results of the baseline and augmentation experiments.% for the non-zero-shot setting. The best performing model is bolded.
} 
\label{table:ST_extrinsic_eval_results}
\end{table*}

\subsection{Effect of Quality on Performance}
\label{sec:quality_performance_performance}
We examine the importance of generating natural CSW sentences in ASR LM rescoring. 
We utilize our human evaluation results from \citet{hamed2023CSAug} and calculate the correlations against ASR scores. The human evaluation involved three annotators assessing 150 sentences across all augmentation techniques for naturalness on a scale of 1 to 5, following the rubrics introduced by \newcite{PC21}. The mean opinion score (MOS) is calculated as the average of the annotators’ scores for each sentence. The percentage of sentences perceived as natural (quite natural but rarely used - perfectly
natural and frequently used) 
per technique is shown in Figure~\ref{fig:human_eval}. %measuring the percentage of synthetic sentences perceived as natural across augmentation techniques and calculate the correlations against ASR scores. 
We report correlations of 0.19 ($p=0.73$) and -0.56 ($p=0.15$) between the zero-shot and non-zero-shot ASR results (presented in Figure~\ref{fig:asr_mt_results}) and the percentage of sentences perceived as natural.
%for naturalness scores against ASR results in zero-shot and non-zero-shot settings.
%between the chrF++ scores (presented in Figure 5) and the percentage of sentences perceived as natural (presented in Figure 2)

%These findings do not hold for MT, 
Unlike ASR, strong positive correlations of 0.92 ($p<0.05$) and 0.97 ($p<0.05$) were reported in zero-shot and non-zero-shot MT settings between chrF++ and naturalness scores. 

To eliminate the factor of varying amounts of generations across techniques, %we perform another set of experiments where we control for this variable. 
we %follow \citet{hamed2023CSAug} and 
conduct constrained experiments (results in Table~\ref{table:ASR_extrinsic_eval_results}), 
%where we restrict the augmentations appended to the baseline training data used in LM rescoring to only those that are successfully augmented across all techniques (= 24.8$k$ sentences). 
where we only utilize the synthetic sentences augmented across all approaches ($=24.8k$ sentences) for LM rescoring. 
%In the constraint experiments, 
We report a correlation of -0.26 ($p=0.54$) between naturalness scores and ASR performance. Therefore, we conclude that for ASR, producing more natural synthetic data does not necessarily entail improvements in ASR LM rescoring.

\begin{table*}[th!]
    \centering
    \begin{tabular}{|l|r|}
    \hline
    \multicolumn{2}{|c|}{\textbf{Examples}}\\\hline
        ASR Ref&
        \<يعني>
        one of my dream jobs
        \<بس ده ده>
        \\
        ASR baseline & 
        \<يعني>
        \textcolor{DarkGreen}{my dream job}
        {\color[HTML]{B42419} \underline{\<انا>}}
        \<بس ده ده >\\
        ASR \BT & 
        \<يعني>
        \textcolor{DarkGreen}{my dream job}
        {\color[HTML]{B42419} \underline{\<برضه>}}
        \<بس ده >
          \\
        MT Ref & \multicolumn{1}{l|}{But this is one of my dream jobs actually.}\\
        MT baseline & \multicolumn{1}{l|}{but this.. \textcolor{DarkGreen}{one of my dream jobs} i mean}\\
        MT \BT & \multicolumn{1}{l|}{but this.. this is \textcolor{DarkGreen}{one of my dream jobs}}\\\hline
        ASR Ref & 
        \<في مصر كلها>
        most beautiful city
        \<اسوان>
        i think
        \<و يعني>
        \\
        ASR baseline & 
        \<في مصر كلها>
        {\color[HTML]{B42419}\underline{ student
         \< ال>}}
        \<اصلا>
        \textcolor{DarkGreen}{i think}
        \<و يعني>\\
        ASR \BT & 
        \<في مصر كلها>
        {\color[HTML]{B42419}\underline{ student
         \< ال>}}
        \textcolor{DarkGreen}{i think}
        \<و يعني>\\
        MT Ref & \multicolumn{1}{l|}{And actually, i think Aswan is the most beautiful city in all Egypt.}\\
        MT baseline & \multicolumn{1}{l|}{and \textcolor{DarkGreen}{i think} aswan \textcolor{DarkGreen}{most beautiful city} in all of egypt}\\
        MT \BT & \multicolumn{1}{l|}{and i mean \textcolor{DarkGreen}{i think} aswan is \textcolor{DarkGreen}{most beautiful city} in all of egypt}\\\hline
        ASR Ref        & 
        \<كتير اوي> 
        posters
        \<بنعمل>
        graphic
        \<ده في ال>
        semester
        \<ال>
        \\
        ASR baseline    & 
        \<كتير اوي>
        {\color[HTML]{B42419} \underline{projects}}
        \<ده في >
        {\color[HTML]{B42419} \underline{\<دراستك>}}
        \<نعمل ال> 
        \textcolor{DarkGreen}{semester}
        \\
        ASR \BT    & 
        \<كتير اوي>
        \textcolor{DarkGreen}{posters}
        \<ده في>
        {\color[HTML]{B42419} \underline{\<دراستك>}}
        \<بنعمل ال>
        \textcolor{DarkGreen}{semester}
        \\
        MT Ref & \multicolumn{1}{l|}{We're designing many posters this semester in graphic.}\\
        MT baseline & \multicolumn{1}{l|}{this \textcolor{DarkGreen}{semester} in \textcolor{DarkGreen}{graphic} we make {\color[HTML]{B42419} \underline{posters a lot}}}\\
        &\multicolumn{1}{p{12cm}|}{(We mark \textit{posters a lot} as incorrect as the output incorrectly follows the same syntactic structure as the original CSW sentence, where the Arabic adjective \<كتير> \textit{ktyr} `many' follows the English noun \textit{posters}.)}\\ 
        MT \BT & \multicolumn{1}{l|}{this \textcolor{DarkGreen}{semester} in \textcolor{DarkGreen}{graphic} we make \textcolor{DarkGreen}{a lot of posters}}\\
        \hline
        %ASR Ref & \<اه انا كنت بلعب> basket\\
        %ASR baseline & \<اه انا كنت بلعب بس كده>\\
        %ASR \BT & \<اه انا كنت بلعب بس كده>\\
        %MT Ref & \multicolumn{1}{l|}{Yes, i used to play basketball.}\\
        %MT baseline & \multicolumn{1}{l|}{yes i was playing \textcolor{DarkGreen}{basket}}\\
        %MT \BT & \multicolumn{1}{l|}{yes i used to play \textcolor{DarkGreen}{basket}}\\
        ASR Ref & 
        %انا بتعلم حاجات كتير جدا منها ان انا ازاي اعمل web applications ازاي اعمل applications ال mobile ازاي اعمل games ازاي حاجات hardware برضه embedded systems
        \<ازاي اعمل >
        web applications
        \<انا بتعلم حاجات كتير جدا منها ان انا ازاي اعمل>\\
        &
        \<برضه>
        hardware
        \< ازاي حاجات>
        games
        \<ازاي اعمل >
        mobile
        \<ال>
        applications
        \\
        &embedded systems
        \\
        ASR baseline & 
        \<ازاي اعمل>
        {\color[HTML]{B42419} \underline{precautions}}
        {\color[HTML]{B42419} \underline{\<ويبقي>}}
        \<و بتعملي حاجات كتير جدا منها ان انا ازاي عامل>\\
         &
         \<برضه>
         \textcolor{DarkGreen}{hardware}
         \<ازاي حاجات>
         \textcolor{DarkGreen}{games}
         \<ازاي اعمل>
         \textcolor{DarkGreen}{mobile}
         \<ال>
         \textcolor{DarkGreen}{applications}\\
         &{\color[HTML]{B42419} \underline{implications}}
         \\
        ASR \BT & 
        \<ازاي اعمل>
        {\color[HTML]{B42419} \underline{precautions}}
        {\color[HTML]{B42419} \underline{\<يبقي>}}
        \<و اتعلمت حاجات كتير جدا منها ان انا ازاي عملوا>\\
        &
        \<برضه>
        \textcolor{DarkGreen}{hardware}
        \<ازاي حاجات>
        \textcolor{DarkGreen}{games}
        \<ازاي اعمل>
        \textcolor{DarkGreen}{mobile}
        \<ال>
        \textcolor{DarkGreen}{applications}\\
        &{\color[HTML]{B42419} \underline{implications}}
        \\
        MT Ref & \multicolumn{1}{p{12cm}|}{I am learning a lot of things, including how to develop web applications, how to develop mobile applications, how to develop games .. as well as hardware things such as embedded systems}\\
        MT baseline & \multicolumn{1}{p{12cm}|}{i learn a lot of things, including how to do  \textcolor{DarkGreen}{web applications}, how to make {\color[HTML]{B42419} \underline{applications the mobile}}, how to make  \textcolor{DarkGreen}{games}, how..  \textcolor{DarkGreen}{hardware embedded systems}}\\
        MT \BT &  \multicolumn{1}{p{12cm}|}{i learn a lot of things, including how to make  \textcolor{DarkGreen}{web applications}, how to make {\color[HTML]{B42419} \underline{applications the mobile}}, how to make  \textcolor{DarkGreen}{games}, how.. \textcolor{DarkGreen}{hardware} also \textcolor{DarkGreen}{embedded systems}}\\\hline
    \end{tabular}
    \caption{Examples of outputs of ASR and MT systems. For each example, we show the reference transcription (ASR Ref) and translation (MT Ref) as well as the outputs of the baseline and \BT~augmentation models in the non-zero-shot setting. 
    The words in the transcriptions/translations are 
    highlighted according to whether they are correct (green) or incorrect (red and underlined) with regards to CSW. Given that Arabic is written from right to left, all sentences are displayed in a right-to-left orientation. 
%We highlight the CSW words that are correctly (green) or incorrectly (red) handled in transcriptions and translations.
}
    \label{table:output_examples}
\end{table*}

\section{Discussion}
\label{sec:discussion}
In this section, we share more insights to gain further understanding of other factors affecting results.

\subsection{Consistency of Results Across Tasks}
%\paragraph{Are findings consistent across tasks?}
We discuss consistency of findings across tasks by comparing our ASR and MT results. With regards to the efficacy of the techniques, we observe that linguistic theories do not show superiority, 
and that the best results are achieved by \BT~followed by \LexPred. The performance of \lexDict~is found to be task-dependent, where it is effective in ASR but not suitable for MT, as the semantics of the original sentences may be altered. 
With regards to the relation between naturalness of generations and NLP improvements, a strong correlation was found for MT, but no correlation for ASR. The importance of quality is also seen in MT, where only \BT~and \LexPred~brought improvements over the baseline in the non-zero-shot setting, as opposed to all approaches in ASR.
%\lexRand~demonstrates high performance, 
%is not necessarily maintained. 
%as it may change the meaning of the original sentence. 
%%, which is confirmed in cascaded ST results. 
%Regarding the naturalness-performance relation, we also find it to be task-dependent, where a strong relation is found for MT but no relation for ASR, where we discuss this in the light of CSW performance across both tasks.
%The dictionary-based replacements, while brings no improvements in MT, provides comparable performance to linguistic theories in ASR. Across both tasks, the linguistic theories do not show superiority and the random lexical replacement demonstrates high performance. Overall, back-translation and \CS~predictive-based lexical replacements consistently provide the best results for both tasks, which was confirmed in the results of the cascaded ST tasks. %However, both approaches are reliant on expensive and limited \CS~parallel data. 

\subsection{Inconsistent Quality-Performance Relation Across Tasks}
%\paragraph{Why is the relation between quality and performance not consistent across tasks?}
We further examine why the relation between quality and performance is not consistent across tasks. 
One factor that may affect this relation is the complexity of the NLP tasks and how well the baseline models perform on CSW. We conduct an error analysis on 100 sentences from ArzEn-ST dev set using the ASR and MT baseline models. We find that 70\% of the sentences in the case of ASR have CSW-related issues as opposed to 25\% in the case of MT. We provide examples in Table~\ref{table:output_examples} demonstrating this disparity in performance. This may be a contributing factor, where quality might be less relevant to low-performing models. While CSW introduces further challenges to ASR, in the case of MT when translating to the primary/secondary language, the translation is partially present in the source sentence, allowing the model to perform better on CSW over monolingual sentences, as shown in \citet{gaser2022exploring}. CSW quality can then be important for the model to not just retain words through translation but to learn the modifications often needed to achieve higher fluency.
%where quality might be less relevant to is more important for well-performing than low-performing models. 
%, but less relevant to low-performing models. 
%this inconsistency between quality and performance has been previously reported in \cite{vaz2024gans}. 
%check that paper for the improtance of quality and diversity across tasks.
%One factor that may affect this relation is the complexity of the NLP tasks and how well the baseline models perform on CSW. It is possible that quality is more important for well-performing than low-performing models. %, but less relevant to low-performing models. For CSW MT, when translating to the primary/secondary language, the translation is partially present in the source sentence, allowing models to achieve better on CSW over monolingual sentences, as shown in \citet{gaser2022exploring}. In this case, CSW quality is important for the model to not just retain words through translation but to learn the modifications often needed to achieve higher fluency. This is not the case in ASR, where CSW complicates the task. We conduct error analysis on 100 sentences from ArzEn-ST dev set using the ASR and MT baseline models, reporting CSW-related issues in 70\% of the sentences in ASR versus 24\% in MT. We provide examples in Appendix~\ref{sec:appendix_examples}. 

\subsection{Other Factors Affecting Performance}
%\paragraph{What other factors in generations can affect ASR performance?} 
We investigate other factors besides quality that may impact the effectiveness of the augmentation techniques, by checking their correlations against MT and ASR non-zero-shot results. With regards to the varying quantity of generated augmentations across techniques, while it may affect results, 
it holds a low correlation of -0.01 ($p=0.98$) and -0.60 ($p=0.12$) with ASR and MT results. We also check correlations against perplexity and OOV rate, where %statistically significant 
strong correlations of 0.89 ($p=0.003$) and 0.84 ($p=0.008$) are found for ASR. %For MT, while a strong correlation -0.72 ($p=0.045$) is also reported for perplexity, no correlation is found for OOV  were found for MT. 
For MT, a lower correlation of -0.77 ($p=0.027$) is found for perplexity (implementation details in Appendix~\ref{sec:mt_ppl}). We do not report correlations with OOV rate for MT, as it is the same value for the majority of augmentation techniques.\footnote{Both Arabic and English sentences of the parallel corpora are used on the source side when training the MT models, so no new words are introduced for \lexRand, \LexPred, and linguistic-based approaches.
%the augmentation approaches relying on injection of words.
}
We agree with \citet{hashimoto2019unifying} that perplexity captures diversity but not quality, while human evaluation captures quality but not diversity, where we believe both criteria affect augmentation performance. 
Accordingly, the high performance achieved by \BT~and \LexPred~across ASR and MT tasks could be supported by their high performance on both criteria.
%We believe both quality and diversity affect augmentation performance. 
%As previously reported in the task of image data augmentation, 
%and as stated in \cite{}, the augmentation performance relies on both. 
%We believe the augmentation performance relies on both; quality as well as diversity. 
%We confirm that the results might be affected however are not solely driven by the quantity of generations, where we report a correlation of -0.38 ($\rho=0.35$) between the WER results in the ASR non-zero shot setting for each augmentation approach and the corresponding number of generated sentences. 

\subsection{Perplexity as a Quality Measure}
%\paragraph{Can perplexity be used as a measure for the quality of CSW generations?}
%We investigate whether perplexity can be used as a measure for the quality of CSW synthetic sentences. 
While perplexity has been previously used to measure the quality of generated CSW and monolingual augmented data %, reflecting their well-formedness 
\cite{WMW+18,feng2020genaug,evuru2024coda},  
%as well as %the fluency/naturalness of 
%monolingual generated data \cite{f},
%we show that it does not correlate with naturalness scores, reporting a correlation of -0.62 ($p=0.10$). 
we report a low correlation of -0.62 ($p=0.10$) with naturalness scores. 
This highlights the importance of assessing naturalness through human evaluations as well as the need for further research towards developing automatic quality evaluation methods %as well as quality metrics ?
for CSW synthetic data.

\section{Conclusions and Outlook}
%This work builds on previous research investigating the efficacy of data augmentation approaches on improvements in CSW MT and the relation between the quality of generated data and the improvements achieved on MT through augmentation We extend previous findings on MT with further results on ASR and cascaded ST to text generalizability of results. While this study extends current literature in CSW data augmentation, further studies are needed to further understand the interplay of data quality and other factors on NLP improvements.
We investigate the efficacy of multiple CSW data augmentation approaches and the relation between quality of generations and improvements. We extend our previous work on MT with results on ASR and ST. %, discussing generalizability of results. %While this study extends current literature, 
We find that back-translation and predictive-based lexical replacements perform consistently well, however, quality of generations are found to be less important for ASR than MT models. % in our experiments. 
We shed light on multiple factors that come into play, %affecting augmentation performance, 
including diversity of generations as well as task complexity and model performance. 

In future work, we plan on expanding the investigated approaches, with a focus on utilizing large language models. We also plan on exploring personalized CSW text generation.
%extending our research to reach a better understanding on the effect of these and other factors. We also plan on expanding the investigated approaches, with a focus on utilizing large language models.

\section*{Limitations}
While this paper provides a comprehensive comparison of CSW augmentation techniques, in terms of the number of augmentation methods and the range of NLP tasks considered, we acknowledge that the coverage is limited to one language pair. Further research is needed to assess the generalizability of our findings across different languages. Additionally, we also acknowledge that LLM-based CSW generation is an interesting direction that is gaining attention \cite{yong2023prompting,potter2024llm,alharbi2024leveraging,kuwanto2024linguistics}. Further research is needed to assess its effectiveness compared to the approaches presented in this work, which we leave for future work. %where we leave this exploration for future work.
%- Only covering one language pair
%- extending it to cover LLMs, especially that the high quality of LLMs hasn't been yet confirmed. Cite existing papers on LLMs' CSW generation.

%, including model performance and task-specific aspects, and the improvements achieved on NLP tasks by the different augmentation techniques.

\section*{Acknowledgments}
We thank Sunayana Sitaram for her insightful discussions. We also thank the reviewers for their valuable comments and constructive feedback.
\bibliography{custom}

\newpage
\onecolumn
\appendix

\section{Statistical Significance Tests}
\label{sec:appendix_full_results}
We present the statistical significance for the ASR and ST experiments in Tables~\ref{table:ASR_stat_sig} and ~\ref{table:ST_stat_sig}.

\begin{table*}[h]
%\begin{subtable*}{}
%\centering
\begin{tabular}{|l|c|lllll|}
\cline{3-7}
\multicolumn{2}{c|}{}& \lexDict & \lexRand & \ecRand & \ecSPF & \mlRand \\\cline{2-7}
\multicolumn{1}{c|}{}& WER & \multicolumn{1}{c}{62.4} & \multicolumn{1}{c}{61.7} & \multicolumn{1}{c}{62.3} & \multicolumn{1}{c}{62.8} & \multicolumn{1}{c|}{62.7} \\\hline
\lexDict  & 62.4 & \cellcolor[HTML]{EFEFEF} & \cellcolor[HTML]{EFEFEF} & \cellcolor[HTML]{EFEFEF} & \cellcolor[HTML]{EFEFEF} & \cellcolor[HTML]{EFEFEF} \\
\lexRand  & 61.7 & 0.009*  & \cellcolor[HTML]{EFEFEF} & \cellcolor[HTML]{EFEFEF} & \cellcolor[HTML]{EFEFEF} & \cellcolor[HTML]{EFEFEF} \\
\ecRand & 62.3 & {\color[HTML]{B42419} 0.719} & 0.017* & \cellcolor[HTML]{EFEFEF} & \cellcolor[HTML]{EFEFEF} & \cellcolor[HTML]{EFEFEF} \\
\ecSPF & 62.8 & {\color[HTML]{B42419} 0.124} & <0.001* & 0.016* & \cellcolor[HTML]{EFEFEF} & \cellcolor[HTML]{EFEFEF} \\
\mlRand & 62.7 & {\color[HTML]{B42419} 0.197} & <0.001* & 0.032* & {\color[HTML]{B42419} 0.719} &\cellcolor[HTML]{EFEFEF} \\
\mlSPF & 62.5 & {\color[HTML]{B42419}0.764} &  0.003* & {\color[HTML]{B42419}0.407} & {\color[HTML]{B42419}0.142}&{\color[HTML]{B42419} 0.254}\\\hline
\multicolumn{7}{c}{}
\end{tabular}
%\caption{Statistical significance between the ASR models in the zero-shot setting. }
%\label{table:ASR_stat_sig_zero-shot}
%\end{subtable*}
%\begin{subtable*}{}
%\centering
\begin{tabular}{|l|c|lllllll|}
\cline{3-9}
\multicolumn{2}{c|}{}& \lexDict& \lexRand& \LexPred& \ecRand& \ecSPF& \mlRand& \mlSPF\\\cline{2-9}
						
\multicolumn{1}{c|}{}& \multicolumn{1}{c|}{WER} & \multicolumn{1}{c}{33.1} & \multicolumn{1}{c}{32.5}& \multicolumn{1}{c}{32.5}& \multicolumn{1}{c}{32.6} & \multicolumn{1}{c}{32.9} & \multicolumn{1}{c}{33.1} & \multicolumn{1}{c|}{33.0}\\\hline
\lexDict  & 33.1 & \cellcolor[HTML]{EFEFEF} & \cellcolor[HTML]{EFEFEF}& \cellcolor[HTML]{EFEFEF} & \cellcolor[HTML]{EFEFEF} & \cellcolor[HTML]{EFEFEF}& \cellcolor[HTML]{EFEFEF}& \cellcolor[HTML]{EFEFEF} \\
\lexRand & 32.5 & 0.006* & \cellcolor[HTML]{EFEFEF}& \cellcolor[HTML]{EFEFEF} & \cellcolor[HTML]{EFEFEF} & \cellcolor[HTML]{EFEFEF}& \cellcolor[HTML]{EFEFEF}& \cellcolor[HTML]{EFEFEF} \\
\LexPred & 32.5 & 0.009* & {\color[HTML]{B42419} 0.928} & \cellcolor[HTML]{EFEFEF} & \cellcolor[HTML]{EFEFEF} & \cellcolor[HTML]{EFEFEF}& \cellcolor[HTML]{EFEFEF} & \cellcolor[HTML]{EFEFEF} \\
\ecRand & 32.6 & {\color[HTML]{B42419} 0.057} & {\color[HTML]{B42419} 0.503} & {\color[HTML]{B42419} 0.535} & \cellcolor[HTML]{EFEFEF} & \cellcolor[HTML]{EFEFEF}& \cellcolor[HTML]{EFEFEF}& \cellcolor[HTML]{EFEFEF} \\
\ecSPF & 32.9 & {\color[HTML]{B42419} 0.337} & {\color[HTML]{B42419} 0.095} & {\color[HTML]{B42419} 0.114} & {\color[HTML]{B42419} 0.238} & \cellcolor[HTML]{EFEFEF}& \cellcolor[HTML]{EFEFEF}& \cellcolor[HTML]{EFEFEF} \\
\mlRand & 33.1 & {\color[HTML]{B42419} 0.865} & 0.003* & 0.004* & 0.018* & {\color[HTML]{B42419} 0.201} & \cellcolor[HTML]{EFEFEF}& \cellcolor[HTML]{EFEFEF} \\
\mlSPF& 33.0 & {\color[HTML]{B42419} 0.667} & 0.020* & 0.026* & {\color[HTML]{B42419} 0.075} & {\color[HTML]{B42419} 0.516} & {\color[HTML]{B42419} 0.465} & \cellcolor[HTML]{EFEFEF} \\
\BT & 32.4 & 0.003* & {\color[HTML]{B42419} 0.589} & {\color[HTML]{B42419} 0.509} & {\color[HTML]{B42419} 0.177} & 0.018* & <0.001* & 0.003* \\\hline
\end{tabular}
%\caption{Statistical significance between the ASR models in the non-zero-shot setting.}
%\label{table:ASR_stat_sig_non-zero-shot}
%\end{subtable*}
\caption{Statistical significance between ASR models in the zero-shot (upper) and non-zero-shot (lower) settings calculated on WER achieved on ArzEn-ST test set \CS~sentences. We present the \textit{p}-values and mark \textit{p}-values < 0.05 with $\ast$, where the null hypothesis can be rejected. We include the WER figures for easier readability and comparison.}
\label{table:ASR_stat_sig}
\end{table*}

\begin{table*}[h]
\centering
\resizebox{\textwidth}{!}{
\begin{tabular}{|l|c|llllllll|}
\cline{3-10}
\multicolumn{2}{c|}{}& \lexDict& \lexRand& \LexPred& \ecRand& \ecSPF& \mlRand& \mlSPF& \BT\\\cline{2-10}
\multicolumn{1}{c|}{}& \multicolumn{1}{c|}{chrF++} & \multicolumn{1}{c}{41.2} & \multicolumn{1}{c}{42.0}& \multicolumn{1}{c}{43.0}& \multicolumn{1}{c}{41.7} & \multicolumn{1}{c}{42.2} & \multicolumn{1}{c}{41.8} & \multicolumn{1}{c}{41.8} & \multicolumn{1}{c|}{43.3} \\\hline
\lexDict  & 41.2 & \cellcolor[HTML]{EFEFEF} & \cellcolor[HTML]{EFEFEF}& \cellcolor[HTML]{EFEFEF} & \cellcolor[HTML]{EFEFEF} & \cellcolor[HTML]{EFEFEF}& \cellcolor[HTML]{EFEFEF}& \cellcolor[HTML]{EFEFEF} & \cellcolor[HTML]{EFEFEF}\\
\lexRand & 42.0 & 0.0010* & \cellcolor[HTML]{EFEFEF}& \cellcolor[HTML]{EFEFEF} & \cellcolor[HTML]{EFEFEF} & \cellcolor[HTML]{EFEFEF}& \cellcolor[HTML]{EFEFEF}& \cellcolor[HTML]{EFEFEF} & \cellcolor[HTML]{EFEFEF}\\
\LexPred & 43.0 & 0.0010* & 0.0010* & \cellcolor[HTML]{EFEFEF} & \cellcolor[HTML]{EFEFEF} & \cellcolor[HTML]{EFEFEF}& \cellcolor[HTML]{EFEFEF} & \cellcolor[HTML]{EFEFEF} & \cellcolor[HTML]{EFEFEF}\\
\ecRand & 41.7 & 0.0100* & 0.0490* & 0.0010* & \cellcolor[HTML]{EFEFEF} & \cellcolor[HTML]{EFEFEF}& \cellcolor[HTML]{EFEFEF}& \cellcolor[HTML]{EFEFEF} & \cellcolor[HTML]{EFEFEF}\\
\ecSPF & 42.2 & 0.0010* & {\color[HTML]{B42419} 0.1598} & 0.0010* & 0.0070* & \cellcolor[HTML]{EFEFEF}& \cellcolor[HTML]{EFEFEF}& \cellcolor[HTML]{EFEFEF} & \cellcolor[HTML]{EFEFEF}\\
\mlRand & 41.8 & 0.0040* & {\color[HTML]{B42419} 0.0939} & 0.0010* & {\color[HTML]{B42419} 0.2687} & 0.0170* & \cellcolor[HTML]{EFEFEF}& \cellcolor[HTML]{EFEFEF} & \cellcolor[HTML]{EFEFEF}\\
\mlSPF& 41.8 & 0.0020* & {\color[HTML]{B42419} 0.1489} & 0.0010* & {\color[HTML]{B42419} 0.1798} & 0.0420* & {\color[HTML]{B42419} 0.2647} & \cellcolor[HTML]{EFEFEF} & \cellcolor[HTML]{EFEFEF}\\
\BT & 43.3 & 0.0010* & 0.0010* & 0.0410* & 0.0010* & 0.0010* & 0.0010* & 0.0010* & \cellcolor[HTML]{EFEFEF}\\\hline\hline
ST\_BL$_{All}$ & 41.6 & 0.0300* & 0.0250* & 0.0010* & {\color[HTML]{B42419} 0.2038} & 0.0040* & {\color[HTML]{B42419} 0.1518} & {\color[HTML]{B42419} 0.0949} & 0.0010*\\\hline
\end{tabular}
}
\label{table:ST_stat_sig_non-zero-shot}
\caption{Statistical significance between ST models in the non-zero-shot setting calculated on the chrF++ scores achieved on ArzEn-ST test set \CS~sentences. We present the \textit{p}-values and mark \textit{p}-values < 0.05 with $\ast$, where the null hypothesis can be rejected. We include the chrF++ scores for easier readability and comparison.}
\label{table:ST_stat_sig}
\end{table*}

\section{Perplexity in MT Setup}
\label{sec:mt_ppl}
We report PPL in MT setups by training transformer-based LMs using Fairseq. The models
are optimized with Adam \cite{kingma2014adam}
using $\beta1$ = 0.9, $\beta2$ = 0.98. We set the dropout to 0.1 and the learning rate to 0.0005. 
We report perplexity for the non-zero-shot settings as follows: \lexDict~(163.2), \lexRand~(156.1), \LexPred~(148.6), \ecRand~(146.0), \ecSPF~(150.5), \mlRand~(147.0), \mlSPF~(150.5), and \BT~(143.2).

%/projekte/slu/Mitarbeiter/hamediy/venv_MT/fairseq/fairseq_cli/train.py --task language_modeling $DATA_DIR --save-dir ${OUT_DIR} --arch transformer_lm --share-decoder-input-output-embed --dropout 0.1 --optimizer adam --adam-betas '(0.9, 0.98)' --weight-decay 0.01 --clip-norm 0.0 --lr 0.0005 --lr-scheduler inverse_sqrt --warmup-updates 4000 --warmup-init-lr 1e-07 --tokens-per-sample 512 --sample-break-mode none --max-tokens 2048 --update-freq 16 --fp16  --max-update 50000 --patience 10
\end{document}